\documentclass[10pt,twocolumn,letterpaper]{article}

\usepackage{cvpr}              

\usepackage{graphicx}
\usepackage{amsmath}
\usepackage{amssymb}
\usepackage{booktabs}
\usepackage{xcolor,colortbl}
\usepackage{multirow}
\usepackage{algorithm,algorithmicx,algpseudocode}
\newcommand{\by}{\mathbf{y}}
\newcommand{\bx}{\mathbf{x}}

\definecolor{Gray}{gray}{0.92}
\definecolor{LightCyan}{rgb}{0.92,1,1}
\definecolor{DarkCyan}{rgb}{0.82,1,1}

\newcolumntype{a}{>{\columncolor{Gray}}c}
\newcolumntype{b}{>{\columncolor{LightCyan}}c}

\usepackage[pagebackref,breaklinks,colorlinks]{hyperref}

\usepackage[capitalize]{cleveref}
\crefname{section}{Sec.}{Secs.}
\Crefname{section}{Section}{Sections}
\Crefname{table}{Table}{Tables}
\crefname{table}{Tab.}{Tabs.}

\def\cvprPaperID{*****} 
\def\confName{CVPR}
\def\confYear{2022}

\begin{document}

\title{Stochastic Trajectory Prediction via Motion Indeterminacy Diffusion}

\author{%
Tianpei Gu\thanks{Equal contribution. \textsuperscript{\dag}Corresponding author.} $  ^{,1,5}$, Guangyi Chen$^{*,2,3}$, Junlong Li$^{4}$, Chunze Lin$^{5}$, Yongming Rao$^{4}$, Jie Zhou$^{4}$, Jiwen Lu$^{\dag,4}$\\
{$^1$University of California, Los Angeles},
{$^2$MBZUAI},
{$^3$Carnegie Mellon University},\\
{$^4$Tsinghua University},
{$^5$SenseTime Research}\\
}
\maketitle

\begin{abstract}
    Human behavior has the nature of indeterminacy, which requires the pedestrian trajectory prediction system to model the multi-modality of future motion states. Unlike existing stochastic trajectory prediction methods which usually use a latent variable to represent multi-modality, we explicitly simulate the process of human motion variation from indeterminate to determinate.
    In this paper, we present a new framework to formulate the trajectory prediction task as a reverse process of motion indeterminacy diffusion (\textbf{MID}), in which we progressively discard indeterminacy from all the walkable areas until reaching the desired trajectory.
    This process is learned with a parameterized Markov chain conditioned by the observed trajectories. We can adjust the length of the chain to control the degree of indeterminacy and balance the diversity and determinacy of the predictions.
    Specifically, we encode the history behavior information and the social interactions as a state embedding and devise a Transformer-based diffusion model to capture the temporal dependencies of trajectories.
    Extensive experiments on the human trajectory prediction benchmarks including the Stanford Drone and ETH/UCY datasets demonstrate the superiority of our method. Code is available at \url{https://github.com/gutianpei/MID}.
\end{abstract}

\section{Introduction}
\label{sec:intro}
Human trajectory prediction plays a crucial role in human-robot interaction systems such as self-driving vehicles and social robots, since human is omnipresent in their environments.
Although significant progresses have been achieved over past few years~\cite{salzmann2020trajectron++,mangalam2020not, YuMa2020Spatio,dendorfer2021mg, mangalam2021goals,Pang_2021_CVPR,sun2021three,zhao2021you}, predicting the future trajectories of pedestrians remains challenging due to the multi-modality of human motion.

\begin{figure}[t]
	\centering
	\includegraphics[width=0.99\linewidth]{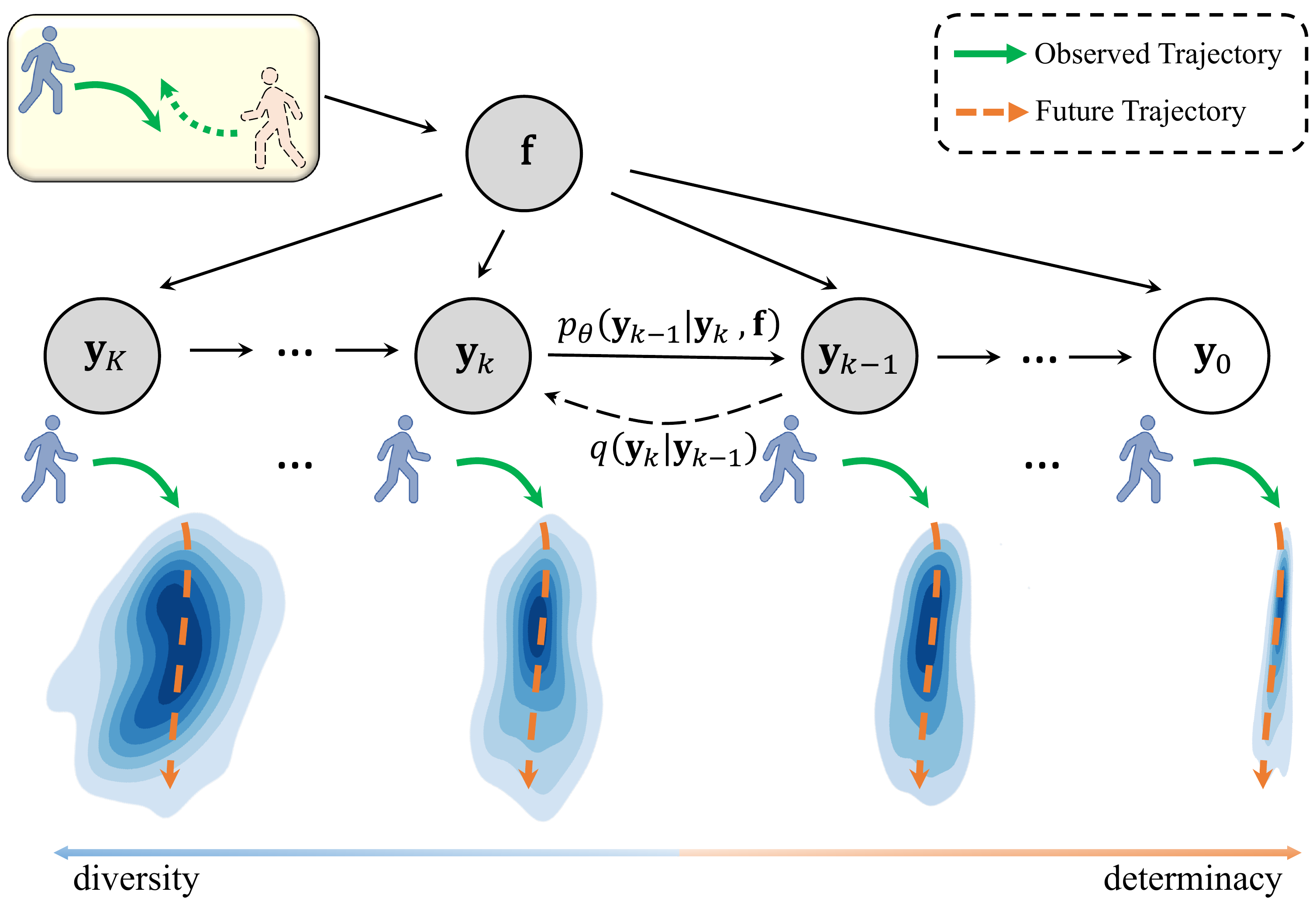}
	\caption{Illustration of the reverse diffusion process for human motion indeterminacy variation. Under high indeterminacy, the trajectory distribution can be regarded as a noise Gaussian distribution which denotes ambiguous walkable areas. With the decreasing of indeterminacy, this distribution gradually approximates the real data distribution to generate desired trajectory. This process from high indeterminacy to low indeterminacy is defined as a reverse diffusion process, in which we learn a Markov chain to progressively discard the indeterminacy. By adjusting the length of the chain, we can make a trade-off between diversity and determinacy, where the longer chain leads to lower diversity and higher determinacy. Best viewed in color.}
	\label{fig:motivation}
	\vspace{-0.5cm}
\end{figure}

The future trajectories of pedestrians are full of indeterminacy, because human can change future motion according to their will or adjust their movement direction based on the surroundings.
Given a history of observed trajectories, there exist many plausible paths that pedestrians could move in the future.
Facing this challenge, most of prior researches apply the generative model to represent multi-modality by a latent variable. For instance, some methods~\cite{gupta2018social,Fang_2020_CVPR,kosaraju2019social,Sun_2020_CVPR,sadeghian2019sophie,zhao2019multi,dendorfer2021mg} utilize generative adversarial networks (GANs) to spread the distribution over all possible future trajectories, while other methods~\cite{salzmann2020trajectron++,ivanovic2019trajectron,lee2017desire,Chen_2021_ICCV,tang2019multiple,Liu_2021_ICCV} exploit conditional variational auto-encoder (CVAE) to encode the multi-modal distribution of future trajectories.
Despite the remarkable progress, these methods still face inherent limitations, \textit{e.g.}, training process could be unstable for GANs due to adversarial learning, and CVAE tends to produce unnatural trajectories.

In this paper, we propose a new trajectory prediction framework, called motion indeterminacy diffusion (MID), to model the indeterminacy of human behavior.
Inspired by non-equilibrium thermodynamics, we consider the future positions as particles in thermodynamics in our framework. The particles (positions) gather and deform to a clear trajectory under low indeterminacy, while stochastically spread over all walkable areas under high indeterminacy. The process of particles evolving from low indeterminacy to high indeterminacy is defined as the diffusion process. This process can be simulated by gradually adding noise to the trajectory until the path is corrupted as Gaussian noise.
The goal of our MID is to reverse this diffusion process by progressively discarding indeterminacy, and converting the ambiguous prediction regions into a deterministic trajectory.
We illustrate the reverse diffusion process of motion indeterminacy in Figure~\ref{fig:motivation}.
Contrary to other stochastic prediction methods that add a noise latent variable on the trajectory feature to obtain indeterminacy, we explicitly simulate the motion indeterminacy variation process.
Our MID learns a Markov chain with parameterized Gaussian transition to model this reverse diffusion process and train it using variational inference conditioned on the observed trajectories.
By choosing different lengths of the chain, we can obtain the predictions with a flexible indeterminacy that is capable of adapting to dynamic environment.
Moreover, our method is more efficient to train than GANs, and is capable of producing more high-quality samples than CVAEs.

To be more specific, we encode the history human trajectories and the social interactions as state embedding via a spatial-temporal graph network. Then, we exploit this state embedding as condition in the Markov chain to guide the learning of reverse diffusion process. To model the temporal dependencies in trajectories, we carefully design a Transformer-based architecture as the core network of MID framework. In the training process, we optimize the model with the variational lower bound, and during the inference, we sample the reasonable trajectories by progressive denoising from a noise distribution.
Extensive experiments demonstrate that our method accurately forecasts reasonable future trajectories with multi-modality, achieving state-of-the-art results on Stanford Drone and ETH/UCY datasets.
We summarize the main contributions of this paper as follows:
\begin{itemize}
\setlength{\itemsep}{2pt}
\setlength{\parsep}{2pt}
\setlength{\parskip}{2pt}
    \item We present a new stochastic trajectory prediction framework with motion indeterminacy diffusion, which gradually discards the indeterminacy to obtain desired trajectory from ambiguous walkable areas.
    \item We devise a Transformer-based architecture for the proposed framework to capture the temporal dependencies in trajectories.
    \item The proposed method achieves state-of-the-art performance on widely used human trajectory prediction benchmarks and provides a potential direction for balancing the diversity and accuracy of predictions.
\end{itemize}

\section{Related Work}
\label{sec:rel_work}
\textbf{Pedestrian Trajectory Prediction:}
Given the observed paths, human trajectory forecasting system aims to estimate the future positions.
Most existing methods formulate trajectory forecasting as a sequential prediction problem and focus on modeling the complex social interaction. For instance, Social Forces~\cite{helbing1995social} introduces attractive and repulsive forces to model human interaction.
With the success of deep learning, many methods design ingenious networks to model the social interactions.
For example, Social-LSTM~\cite{alahi2016social} devises a social pooling layer to aggregate the interaction information of neighborhoods.
Some methods apply the attention models~\cite{fernando2018soft+,vemula2018social,sadeghian2019sophie,zhang2019sr,kosaraju2019social} to explore the key interactions of the crowd.
In addition, the spatial-temporal graph model is applied to jointly model the temporal clues and social interactions~\cite{huang2019stgat,ivanovic2019trajectron,mohamed2020social,Sun_2020_CVPR2,yu2020spatio,salzmann2020trajectron++}.
Beyond social interactions, many methods incorporate the physical environment interactions by introducing the map images~\cite{sadeghian2019sophie,lee2017desire,kosaraju2019social,mangalam2021goals,dendorfer2021mg}.
Recently, some methods analyze the effect of social interaction and show it is biased~\cite{chen2021human,makansi2021you}.

\textbf{Stochastic Prediction Model:}
Due to the inherent indeterminacy of human behavior, Many stochastic prediction methods are proposed to model the multi-modality of future motions. Some methods~\cite{gupta2018social,Fang_2020_CVPR,kosaraju2019social,Sun_2020_CVPR,sadeghian2019sophie,zhao2019multi,dendorfer2021mg} employ GANs~\cite{goodfellow2014generative} to model the multi-modality with a noise variable, and another line of methods~\cite{salzmann2020trajectron++,ivanovic2019trajectron,lee2017desire,Chen_2021_ICCV,tang2019multiple,Liu_2021_ICCV} apply the CVAE~\cite{sohn2015learning} instead. Besides, some methods~\cite{liang2020simaug,liang2020garden,deo2020trajectory} propose to learn the grid-based location encoder for multi-modal probability prediction. Recently, the goals of pedestrians~\cite{mangalam2021goals,zhao2020tnt,mangalam2020not,zhao2021you} are introduced in the trajectory prediction system as condition to analyze the probability of multiple plausible endpoints.
While remarkable progress have been made, these stochastic prediction methods have some inherent limitations, \textit{e.g.}, the unstable training or unnatural trajectories.
In this paper, we propose a new stochastic framework with motion indeterminacy diffusion, which formulates the trajectory prediction problem as a process from an ambiguous walkable region to the desired trajectory.

\textbf{Denoising Diffusion Probabilistic Models:} Denoising diffusion probabilistic models (DDPM)~\cite{ho2020denoising, sohl2015deep}, as known as diffusion models for brevity, are a class of deep generative models inspired by  non-equilibrium thermodynamics. It is first proposed by Sohl-Dickstein \emph{et al.}~\cite{sohl2015deep} and attracts much attention recently due to state-of-the-art performance in various generation tasks including image generation ~\cite{ho2020denoising,nichol2021improved,dhariwal2021diffusion, choi2021ilvr}, 3D point cloud generation~\cite{zhou20213d, luo2021diffusion}, and audio generation~\cite{kong2020diffwave,chen2020wavegrad,popov2021grad}. The diffusion models generally learn a parameterized Markov chain to gradually denoise from an original common distribution to a specific data distribution.
In this paper, we introduce the diffusion model to simulate the variation of indeterminacy for trajectory prediction, and design a Transformer-based architecture for the temporal dependency of trajectories.

\begin{figure*}[h]
    \scriptsize
    \setlength{\tabcolsep}{1.5pt}
    \centering
    \includegraphics[width=1\linewidth]{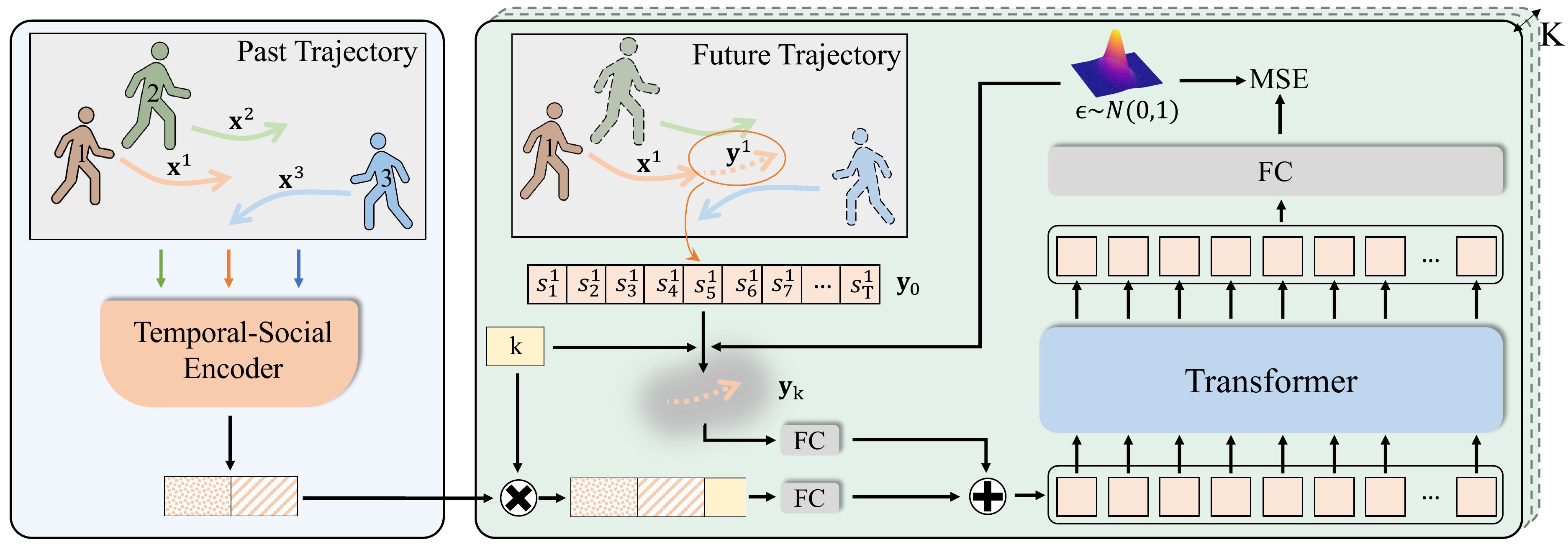}
    \\
    \caption{The architecture of our MID framework. MID consists of a temporal-social encoder network and a Transformer-based decoder network. The encoder maps the history path and social interaction clues into a state embedding. The decoder then takes $\mathbf{y}_{k}$ along with state embedding and the time embedding, where $\mathbf{y}_{k}$ is corrupted $k$ times by a noise variable from the ground truth trajectory $\mathbf{y}_{0}$. We learn the model with the MSE loss between the model output and a noise variable in standard Gaussian distribution. }
    \label{fig:pipeline}
    \vspace{-14pt}
\end{figure*}

\section{Proposed Approach}
\label{sec:method}
In this section, we introduce our MID method, which models stochastic trajectory prediction task by motion indeterminacy diffusion.
We first explicitly formulate the indeterminacy variation as a reverse diffusion process.
Then we describe how to train this diffusion model using the variational inference. Finally, we present the detailed network architecture of our method shown in Figure~\ref{fig:pipeline}.

\subsection{Problem Formulation}
The goal of pedestrian trajectory prediction is to generate plausible future trajectories for pedestrians based on their prior movements.
The input of the prediction system is the $N$ history trajectories in a scene such that $\mathbf{x}^i =\{s^i_t \in \mathbb{R}^2| t=-T_{init},-T_{init}+1, \cdots, 0\} $, $\forall i \in \{1,2,\cdots,N\} $, where the $s^i_t$ is the 2D location at  timestamp $t$, $T_{init}$ denotes the length of the observed trajectory, and the current timestamp is $t=0$. Similarly, the predicted future trajectories can be written as $\mathbf{y}^i =\{s^i_t \in \mathbb{R}^2| t=1,2,\cdots,T_{pred} \}$.
For clarity, we use $\mathbf{x}$ and $\mathbf{y}$ without the superscript $i$ for the history and future trajectory in the following subsections.

\subsection{Motion Indeterminacy Diffusion}
Due to the indeterminacy of human behavior, one person has multiple plausible paths in future state. Thus, we present a new framework to formulate the stochastic trajectory prediction by motion indeterminacy diffusion.
Unlike other stochastic prediction methods that add a latent variable on the trajectory feature to obtain indeterminacy, our MID generates the trajectory by gradually reducing the indeterminacy from all walkable areas to the determinate prediction with a parameterized Markov chain.

As shown in Figure~\ref{fig:motivation}, given the initial ambiguous region $\mathbf{y}_{K}$ under the noise distribution and the desired trajectory $\mathbf{y}_{0}$ under the data distribution, we define the diffusion process as $(\mathbf{y}_{0},\mathbf{y}_{1},\cdots, \mathbf{y}_{K})$, where $K$ is the maximum number of diffusion steps. This process aims to gradually add the indeterminacy until the ground truth trajectory is corrupted into a noisy walkable region. On the contrary, we learn the reverse process as $(\mathbf{y}_{K},\mathbf{y}_{K-1},\cdots, \mathbf{y}_{0})$ to gradually reduce the indeterminacy from $\mathbf{y}_{K}$ to generate the trajectories. Both diffusion and reverse diffusion processes are formulated by a Markov chain with Gaussian transitions.

First, we formulate the posterior distribution of the diffusion process from $\mathbf{y}_{0}$ to $\mathbf{y}_{K}$ as:
\begin{equation}
\setlength{\abovedisplayskip}{1pt}
\setlength{\belowdisplayskip}{1pt}
\begin{aligned}
\label{eq: diffusion_q} q(\mathbf{y}_{1:K}|\mathbf{y}_{0}) &:=  \prod_{k=1}^{K} q(\mathbf{y}_{k}|\mathbf{y}_{k-1}) \\
q(\mathbf{y}_{k}|\mathbf{y}_{k-1}) &:= \mathcal{N}(\mathbf{y}_{k}; \sqrt{1-\beta_{k}}\mathbf{y}_{k-1},\beta_{k}\mathbf{I}),
\end{aligned}
\end{equation}
where $\beta_{1}, \beta_{2}, \cdots \beta_{K}$ are fixed variance schedulers that control the scale of the injected noise. Due to the notable property of the Gaussian transitions, we calculate the diffusion process at any step $k$ in a closed form as:
\begin{equation}
\setlength{\abovedisplayskip}{4pt}
\setlength{\belowdisplayskip}{4pt}
\begin{aligned}
\label{eq: q_yt_by_y0} q(\mathbf{y}_{k}|\mathbf{y}_{0}) :=  \mathcal{N}(\mathbf{y}_{k}; \sqrt{\Bar{\alpha_k}}\mathbf{y}_{0},(1-\Bar{\alpha_k})\mathbf{I}),
\end{aligned}
\end{equation}
where $\alpha_{k} = 1 - \beta_{k}$ and $\Bar{\alpha_{k}} = \prod_{s=1}^{k}\alpha_{s}$. Therefore, when $K$ is large enough, we approximately obtain that $\mathbf{y}_{K} \sim \mathcal{N}(\mathbf{0},\mathbf{I})$.
It indicates that the signal is corrupted into a Gaussian noise distribution when gradually adding noise, which conforms to the non-equilibrium thermodynamics phenomenon of diffusion process.

Next, we formulate the trajectories generation process as a reverse diffusion process from noise distribution.
We model this reverse process by parameterized Gaussian transitions with the observed trajectories as condition.
Given a state feature $\mathbf{f}$ learned by a temporal-social encoder $\mathcal{F_{\psi}}$ parameterized by $\psi$ with the history trajectories $\mathbf{x}$ as input, we formulate the reverse diffusion process as:
\begin{equation}
\setlength{\abovedisplayskip}{1pt}
\setlength{\belowdisplayskip}{1pt}
\begin{aligned}
\label{eq: denoise_p} p_{\theta}(\mathbf{y}_{0:K} | \mathbf{f}) &:=  p(\mathbf{y}_{K})\prod_{k=1}^{K} p_{\theta}(\mathbf{y}_{k-1}|\mathbf{y}_{k}, \mathbf{f}) \\
 p_{\theta}(\mathbf{y}_{k-1}|\mathbf{y}_{k}, \mathbf{f}) &:= \mathcal{N}(\mathbf{y}_{k-1}; \boldsymbol{\mu}_{\theta}(\mathbf{y}_{k},k,\mathbf{f});\boldsymbol{\Sigma}_{\theta}(\mathbf{y}_{k},k)), \\
\end{aligned}
\end{equation}
where $p(\mathbf{y}_{K})$ is an initial noise Gaussian distribution, and $\theta$ denotes the parameter of the diffusion model. Both parameters of diffusion model $\theta$ and encoder network $\psi$ are trained using the trajectory data. Note that we share the network parameters for all transitions. As shown the previous work~\cite{ho2020denoising}, the variance term of the Gaussian transition can be set as $\boldsymbol{\Sigma}_{\theta}(\mathbf{y}_{K},k) = \sigma_{k}^2\mathbf{I} = \beta_{k}\mathbf{I}$. This setting denotes the upper bound on reverse process entropy for data and shows good performance in practice~\cite{sohl2015deep}.

\subsection{Training Objective}

Having formulated diffusion and reverse diffusion processes, we describe how to train the diffusion model.
To predict the real trajectory $\mathbf{y}_{0}$, the desired training should optimize the log-likelihood  $\mathbb{E}[\log{p_{\theta}}(\mathbf{y}_{0})]$ in the reverse process. However, the exact log-likelihood is intractable, we thus maximize the variational lower bound for optimization:
\begin{equation}
\setlength{\abovedisplayskip}{1pt}
\setlength{\belowdisplayskip}{1pt}
\begin{aligned}
\label{eq: ELBO}
\mathbb{E}[\log{p_{\theta}}(\mathbf{y}_{0})] &  \geq  \mathbb{E}_{q}[\log\frac{p_{\theta}(\mathbf{y}_{0:K}, \mathbf{f})}{q(\mathbf{y}_{1:K} \vert \mathbf{y}_{0})}] \\
& = \mathbb{E}_{q}[\log p(\mathbf{y}_{K})  +  \sum_{k=1}^{K}\log\frac{p_{\theta}(\mathbf{y}_{k-1} | \mathbf{y}_{k}, \mathbf{f})}{q(\mathbf{y}_{k}|\mathbf{y}_{k-1})}].
\end{aligned}
\end{equation}
We utilize the negative bound as the loss function and perform the training by optimizing it as:
\begin{equation}
\setlength{\abovedisplayskip}{1pt}
\setlength{\belowdisplayskip}{1pt}
\begin{aligned}
\label{eq: loss} L(\theta,\psi) & =  \mathbb{E}_{q}[\sum_{k=2}^{K}D_{KL}(q(\mathbf{y}_{k-1}\vert \mathbf{y}_{k},\mathbf{y}_{0}) \Vert p_{\theta}(\mathbf{y}_{k-1} \vert \mathbf{y}_{k}, \mathbf{f})) \\
&  -  \log p_{\theta}(\mathbf{y}_{0} \vert \mathbf{y}_{1},\mathbf{f})].
\end{aligned}
\end{equation}
In this loss function, we ignore the term with $\mathbb{E}_{q}\log p(\mathbf{y}_{K})$ in \eqref{eq: ELBO}, since $p(\mathbf{y}_{K})$ is a standard Gaussian and $q(\mathbf{y}_{K}|\mathbf{y}_{0})$ has no learnable parameters as shown in \eqref{eq: q_yt_by_y0}.

Here we describe how to calculate the first term $D_{KL}$. The posterior $q(\mathbf{y}_{k-1}\vert \mathbf{y}_{k},\mathbf{y}_{0})$ in $D_{KL}$ is tractable and can be represented by Gaussian distribution as:
\begin{equation}
\label{eq: q_posterior} q(\mathbf{y}_{k-1}\vert \mathbf{y}_{k},\mathbf{y}_{0}) =
\mathcal{N}(\mathbf{y}_{k-1}; \Tilde{\boldsymbol{\mu}}_{k}(\mathbf{y}_{k},\mathbf{y}_{0}), \Tilde{\beta}_{k}\mathbf{I}),
\end{equation}
 where the closed form of $\Tilde{\boldsymbol{\mu}}_{k}(\mathbf{y}_{k},\mathbf{y}_{0})$ and $\Tilde{\beta}_{k}$ is calculated as:
\begin{equation}
\setlength{\abovedisplayskip}{1pt}
\setlength{\belowdisplayskip}{1pt}
\begin{aligned}
\label{eq: q_mu_sigma}  \Tilde{\boldsymbol{\mu}}_{k}(\mathbf{y}_{k},\mathbf{y}_{0}) &= \frac{\sqrt{\Bar{\alpha}_{k-1}}\beta_{k}}{1-\Bar{\alpha_{k}}}\mathbf{y}_{0} + \frac{\sqrt{\alpha_{k}}(1-\Bar{\alpha}_{k-1})}{1-\Bar{\alpha}_{k}}\mathbf{y}_{k} \cr
\Tilde{\beta}_{k} &= \frac{1-\Bar{\alpha}_{k-1}}{1-\Bar{\alpha}_{k}}\beta_{k}\mathbf{I}.
\end{aligned}
\end{equation}
Since both diffusion process \eqref{eq: q_posterior} and reverse process \eqref{eq: denoise_p} are Gaussian, we can calculate the $D_{KL}$ by the difference between the means of $\Tilde{\boldsymbol{\mu}}_{k}$ and $\boldsymbol{\mu}_{\theta}$ as:
\begin{equation}
\label{eq: kl_simplified}  D_{KL} = \mathbb{E}_{q}\left[ \lambda \Vert \Tilde{\boldsymbol{\mu}}_{k}(\mathbf{y}_{k},\mathbf{y}_{0}) - \boldsymbol{\mu}_{\theta}(\mathbf{y}_{k},k,\mathbf{f})\Vert^{2} \right] +C,
\end{equation}
where $\lambda$ and $C$ are coefficients with no effect on the gradient direction. Note that the second term $- \log p_{\theta}(\mathbf{y}_{0} \vert \mathbf{y}_{1},\mathbf{f}) $ can also be formulated as the form in \eqref{eq: kl_simplified} when $k=1$.
Finally, we apply the parameterization method as shown in ~\cite{ho2020denoising} to reparameterize:
\begin{equation}
\setlength{\abovedisplayskip}{1pt}
\setlength{\belowdisplayskip}{1pt}
\begin{aligned}
\label{eq: reparam_mu} \boldsymbol{\mu}_{\theta}(\mathbf{y}_{k},k,\mathbf{f}) = \frac{1}{\sqrt{\alpha_{k}}}(\mathbf{y}_{k} - \frac{\beta_{k}}{\sqrt{1-\Bar{\alpha_{k}}}}\epsilon_{\theta}(\mathbf{y}_{k},k,\mathbf{f})),
\end{aligned}
\end{equation}
and obtain a simplified loss function as:
\begin{equation}
\begin{aligned}
\label{eq: loss_simple} L(\theta, \psi) = \mathbb{E}_{\epsilon,\mathbf{y}_{0},k} \Vert \epsilon - \epsilon_{(\theta, \psi)}(\mathbf{y}_{k}, k, \mathbf{x})  \Vert,
\end{aligned}
\end{equation}
where $\epsilon \sim \mathcal{N}(0,\mathbf{I})$, $\mathbf{y}_{k} = \sqrt{\Bar{\alpha_{k}}} \mathbf{y}_{0} + \sqrt{1-\Bar{\alpha_{k}}} \epsilon $ and the training is performed at each step $k \in {1,2,\cdots,K}$. (Please see the detailed derivation and detail algorithms in the supplementary material.)

\subsection{Inference}

Once the reverse process is trained, we can generate the plausible trajectories by a noise Gaussian $\mathbf{y}_{K} \sim \mathcal{N}(0, \mathbf{I})$ through the reverse process $p_{\theta}$.
With the reparameterization in \eqref{eq: reparam_mu}, we generate the trajectories from $\mathbf{y}_{K}$ to $\mathbf{y}_{0}$ as:
\begin{equation}
\setlength{\abovedisplayskip}{1pt}
\setlength{\belowdisplayskip}{1pt}
\begin{aligned}
\label{eq: sample} \mathbf{y}_{k-1} =\frac{1}{\sqrt{\alpha_{k}}}(\mathbf{y}_{k} - \frac{\beta_{k}}{\sqrt{1-\Bar{\alpha_{k}}}}\epsilon_{\theta}(\mathbf{y}_{k},k,\mathbf{f})) + \sqrt{\beta_{k}}\mathbf{z},
\end{aligned}
\end{equation}
where $\mathbf{z}$ is a random variable in standard Gaussian distribution and $\epsilon_{\theta}$ is the trained network whose inputs include the previous step's prediction $\mathbf{y}_{k}$, state embedding $\mathbf{f}$, and step $k$.

\subsection{Network Architecture}

Different from the widely used UNet~\cite{ronneberger2015u} in image-based diffusion models~\cite{ho2020denoising,nichol2021improved,dhariwal2021diffusion}, we design a new Transformer-based network architecture for our MID. With the Transformer, the model can better explore the temporal dependency of paths for the trajectory prediction task. To be specific, MID consists of two key networks: an encoder network with parameters $\psi$ which learns the state embedding by observed history trajectories and their social interactions, and a Transformer-based decoder parameterized by $\theta$ for the reverse diffusion process.
An overview of the whole architecture is depicted in Figure~\ref{fig:pipeline}.
We will introduce each part in detail in the following.

The encoder network models the history behaviors and social interactions as the state embedding $\mathbf{f}$. This embedding is fed into the decoder network as the condition of the diffusion model. Note that, designing the network to model social interactions is not the main focus of this work, and MID is an encoder-agnostic framework which can directly equip with different encoders introduced in previous methods. In the experiments, we apply the encoder of Trajectron++~\cite{salzmann2020trajectron++} for its superior representation ability.

For the decoder, we design a Transformer-based architecture to model the Gaussian transitions in Markov chain. As shown in Figure~\ref{fig:pipeline}, the inputs of decoder include the ground truth trajectory $\mathbf{y}_{0}$, the noise variable $\epsilon \sim \mathcal{N}(\mathbf{0},\mathbf{I})$, the condition feature $\mathbf{f}$ from the encoder, and a time embedding. In step $k$, we first add noise into trajectory to get $\mathbf{y}_{k} = \sqrt{\Bar{\alpha_{k}}} \mathbf{y}_{0} + \sqrt{1-\Bar{\alpha_{k}}} \epsilon $. Simultaneously, we calculate the time embedding and concatenate it with the feature of observed trajectory. Then, we apply fully-connected layers to upsample both trajectory $\mathbf{y}_{k}$ and condition $\mathbf{f}$, then sum up the outputs as the fused feature. We also introduce the positional embedding in the form of sinusoidal functions on the summation to emphasize the positional relation at different trajectory timestamp $t$. Finally, the fused feature with positional embedding is fed into the Transformer network to learn the complex spatial-temporal clues. The Transformer-based decoder network consists of three self-attention layers to sufficiently model the temporal dependencies in trajectories, which takes the high dimension sequence as input and outputs the sequence with the same dimension. With a fully-connected layer, we downsample the output sequence to the trajectory dimension.  We finally perform mean square error (MSE) loss between the output and a random Gaussian as \eqref{eq: loss_simple} for current iteration to optimize the network. Please see the network details in the supplementary material.

\section{Experiments}
In this section, we first compared the proposed method with state-of-the-art approaches on two widely-used pedestrian trajectories prediction benchmarks, then conducted ablation studies to analyze the effectiveness of key components of our MID framework and provided an analysis regarding the reverse diffusion process.

\subsection{Experimental Setup}

\textbf{Datasets:}
We evaluated our method on two public pedestrian trajectories forecasting benchmarks including Stanford Drone Dataset (SDD)~\cite{robicquet2016learning} and UCY/ETH~\cite{pellegrini2010improving,lerner2007crowds}.

\textit{Stanford Drone Dataset:} Stanford Drone Dataset~\cite{robicquet2016learning} is a well established benchmark for human trajectory prediction in bird’s eye view. The dataset consists of 20 scenes captured using a drone in top down view around the university campus containing several moving agents like humans and vehicles.

\textit{ETH/UCY:} The ETH~\cite{pellegrini2010improving} and UCY~\cite{lerner2007crowds} dataset group consists of five different scenes – ETH \& HOTEL (from ETH) and UNIV, ZARA1, \& ZARA2 (from UCY). All the scenes report the position of pedestrians in world-coordinates and hence the results we report are in metres. The scenes are captured in unconstrained environments with few objects blocking pedestrian paths.

\begin{table}[t]
\renewcommand\arraystretch{1.1}
\renewcommand\tabcolsep{6pt}
\newcolumntype{g}{>{\columncolor{Gray}}c}
    \centering
    \normalsize
    \caption{Quantitative results on the Stanford Drone dataset with Best-of-20 strategy in ADE/FDE metric. ``T'' denotes the method only using the trajectory position information, and `T + I'' denotes the method using both position and visual image information.
    $\dagger$ means the results are reproduced by us with the official released code. Lower is better.}
    \label{tab: sdd}
    \small
    \begin{tabular}{c | c | c |c|g }
    \hline
    \hline
     \textbf{Methods} & \textbf{Input}  &\textbf{Sampling}  &\textbf{ADE} & \textbf{FDE} \\
    \hline
      CGNS~\cite{li2019conditional} & T + I &20& 15.60&28.20\\
      SimAug~\cite{liang2020simaug}&T + I &20 &10.27  & 19.71\\
      $\dagger$Y-Net~\cite{mangalam2021goals}  & T + I&20&8.97&14.61\\
     Y-Net~\cite{mangalam2021goals}+ TTST & T + I&10000&7.85& {11.85}\\
      \hline
      Social-GAN~\cite{gupta2018social}& T &20& 27.23 & 41.44\\
      PECNet~\cite{mangalam2020not}& T&20&9.96   & 15.88\\
      $\dagger$Trajectron++~\cite{salzmann2020trajectron++} &T &20 &8.98  & 19.02\\
      LB-EBM~\cite{Pang_2021_CVPR}&T &20&8.87  & 15.61\\
      PCCSNET~\cite{sun2021three}& T &20 &8.62 & 16.16 \\
      $\dagger$Expert~\cite{zhao2021you}&T &20 & 10.67 & 14.38 \\
      $\dagger$ Expert~\cite{zhao2021you}+GMM&T &20$\times$20& 7.65 & 14.38 \\
    \hline
      \textbf{MID}& T&20 & {7.61} & 14.30\\
    \hline
    \hline
\end{tabular}
\label{table:sdd}
\vspace{-0.3cm}
\end{table}

\begin{table*}
\caption{Quantitative results on the ETH/UCY dataset with Best-of-20 strategy in ADE/FDE metric. Lower is better.}
\vspace{-0.4cm}
\linespread{3.0}
\renewcommand\arraystretch{1.1}
\renewcommand\tabcolsep{3pt}
\begin{center}
\newcolumntype{g}{>{\columncolor{Gray}}c}
\newcolumntype{y}{>{\columncolor{LightCyan}}c}
\newcolumntype{d}{>{\columncolor{DarkCyan}}c}
\begin{tabular}{l| c| c|c g c g c g c g c g y d}
\hline
\hline
~& \multirow{2}{*}{\textbf{Input}} &\multirow{2}{*}{\textbf{Sampling}} &\multicolumn{2}{c}{\textbf{ETH}} & \multicolumn{2}{c}{\textbf{HOTEL}} &  \multicolumn{2}{c}{\textbf{UNIV}} &  \multicolumn{2}{c}{\textbf{ZARA1}} & \multicolumn{2}{c}{\textbf{ZARA2}} & \multicolumn{2}{c}{\textbf{AVG}} \\
\cline{4-15}
~ &~ & &\multicolumn{1}{c}{\textbf{ADE}} & \multicolumn{1}{c}{\textbf{FDE}} &  \multicolumn{1}{c}{\textbf{ADE}} &  \multicolumn{1}{c}{\textbf{FDE}} &
\multicolumn{1}{c}{\textbf{ADE}} &  \multicolumn{1}{c}{\textbf{FDE}} &
\multicolumn{1}{c}{\textbf{ADE}} &  \multicolumn{1}{c}{\textbf{FDE}} &
\multicolumn{1}{c}{\textbf{ADE}} &  \multicolumn{1}{c}{\textbf{FDE}} &
\multicolumn{1}{c}{\textbf{ADE}} &  \multicolumn{1}{c}{\textbf{FDE}}

\\
\hline
SoPhie~\cite{sadeghian2019sophie}&T + I&20 &0.70&1.43 & 0.76&1.67 & 0.54&1.24  &0.30&0.63   & 0.38&0.78 & 0.54&1.15\\
CGNS~\cite{li2019conditional}&T + I&20 &0.62&1.40&0.70&0.93&0.48&1.22&0.32&0.59&0.35&0.71&0.49&0.97 \\
Social-BiGAT~\cite{kosaraju2019social} & T + I &20 &0.69&1.29  & 0.49&1.01& 0.55&1.32 & 0.30&0.62   &0.36&0.75  & 0.48&1.00\\
MG-GAN~\cite{dendorfer2021mg} &T + I&20 &0.47&0.91  & 0.14&0.24 & 0.54&1.07 & 0.36&0.73  & 0.29&0.60 & 0.36&0.71 \\
Y-Net~\cite{mangalam2021goals} + TTST &T + I&10000 & 0.28&{0.33} & {0.10}&{0.14}&0.24&0.41  &  0.17&{0.27} &{0.13}&{0.22} &0.18&{0.27} \\
\hline
Social-GAN~\cite{gupta2018social} &T&20 &0.81&1.52   &0.72&1.61 &0.60&1.26 &0.34&0.69  & 0.42&0.84  &0.58&1.18  \\
Causal-STGCNN~\cite{chen2021human}& T&20 &0.64&1.00  & 0.38&0.45 & 0.49&0.81 & 0.34&0.53  & 0.32&0.49 & 0.43&0.66 \\
PECNet~\cite{mangalam2020not} &T&20 & 0.54&0.87 & 0.18&0.24 &0.35&0.60  & 0.22&0.39  & 0.17&0.30 &0.29&0.48 \\
STAR~\cite{YuMa2020Spatio} & T &20 &0.36&0.65&0.17&0.36&0.31&0.62&0.26&0.55&0.22&0.46&0.26&0.53\\
Trajectron++~\cite{salzmann2020trajectron++} &T &20 & 0.39&0.83  & 0.12&0.21& 0.20&0.44 & 0.15&0.33  &0.11&0.25  &0.19&0.41 \\
LB-EBM~\cite{Pang_2021_CVPR} &T&20 & 0.30&0.52 & 0.13&0.20 &0.27&0.52  & 0.20&0.37  & 0.15&0.29 &0.21&0.38 \\
PCCSNET~\cite{sun2021three} &T&20 & 0.28 & 0.54 & 0.11 &0.19&0.29&0.60&0.21&0.44&0.15&0.34&0.21&0.42 \\
$\dagger$Expert~\cite{zhao2021you} &T&20 &  0.37&0.65&0.11&0.15&0.20&0.44&0.15&0.31&0.12&0.26&0.19&0.36\\
$\dagger$Expert~\cite{zhao2021you}+GMM &T&20$\times$20 &  0.29&0.65&{0.08}&{0.15}&0.15&0.44&0.11&0.31& 0.09&0.26&0.14&0.36\\
\hline
\textbf{MID} & T &20 & 0.39& 0.66 & 0.13& 0.22&0.22&0.45  &  0.17&0.30 &0.13&0.27 &0.21&0.38  \\
\hline\hline
\end{tabular}
\end{center}
\label{table:eth_ucy}
\vspace{-0.6cm}
\end{table*}

\textbf{Evaluation Metric:}
We adopted the widely-used evaluation metrics Average Displacement Error (ADE) and Final Displacement Error (FDE).
ADE computes the average error between all the ground truth positions and the estimated positions in the trajectory, and FDE computes the displacement between the end points of ground truth and predicted trajectories.
The trajectories are sampled 0.4 seconds interval, where the first 3.2 seconds of a trajectory is used as observed data to predict the next 4.8 seconds future trajectory. For the  ETH/UCY dataset, we followed the leave one out cross-validation evaluation strategy such that we trained our model on four scenes and tested on the remaining one~\cite{gupta2018social,kosaraju2019social,huang2019stgat,salzmann2020trajectron++}.
Considering the stochastic property of our method, we used Best-of-N strategy to compute the final ADE and FDE with $N = 20$.

\textbf{Implementation Details:}
We devised a three-layers Transformer as the core network for our MID, where the Transformer dimension is set to 512, and 4 attention heads are applied.
We employed one fully-connected layer to upsample the input of the model from dimension 2 to the Transformer dimension, and another fully-connected layer to upsample the observed trajectory feature $\mathbf{f}$ to the same dimension.
We utilized three fully-connected layers to progressively downsample the Transformer output sequence to the predicted trajectory, such that $512$d-$256$d-$2$d.
The training was performed with Adam optimizer, with a learning rate of $0.001$ and batch size of $256$.
All the experiments were conducted on a single Tesla V100 GPU.

\begin{table}
\caption{Ablation studies on MID and network architecture designs. Trans is the abbreviation of Transformer}
\vspace{-0.4cm}
\linespread{3.0}
\renewcommand\arraystretch{1.1}
\renewcommand\tabcolsep{3.7pt}
\begin{center}
\newcolumntype{g}{>{\columncolor{Gray}}c}
\newcolumntype{y}{>{\columncolor{LightCyan}}c}
\newcolumntype{d}{>{\columncolor{DarkCyan}}c}

\begin{tabular}{c | c | c |c|g }
\hline
\hline
 \textbf{Group} & \textbf{Method}  &\textbf{Architecture}  &\textbf{ADE} & \textbf{FDE} \\
    \hline
1&MID & Trans-512d  & 7.61&14.30\\
\hline
\multirow{2}{*}{2}&MID & Trans-256d & 7.91&14.50\\
~&MID & Trans-1024d & 7.64&14.37\\
\hline
\multirow{2}{*}{3}&MID & Linear & 8.85&17.25\\
~&MID & LSTM & 8.41&16.57\\
\hline
\multirow{2}{*}{4}&Trajectron++& LSTM & 8.98 & 19.02 \\
~&Trajectron++ & Trans-256d & 9.86 &19.56\\
\hline\hline
\end{tabular}
\end{center}
\label{table:ablation_ddpm}
\vspace{-0.6cm}
\end{table}

\subsection{Comparison with state-of-the-art methods}
We quantitatively compare our method with a wide range of current methods.
As shown in Table~\ref{table:sdd}, we provide the comparison between our method and existing methods on the Stanford Drone dataset. We categorize methods as Trajectory-Only methods (T) and Trajectory-and-Image (T+I) methods, as the additional image information may be crucial at certain circumstances yet increases computation cost. Besides, we also report the sampling number since adding the sampling number can effectively promote the performance. We provide the results under standard 20 samplings of MID and other methods for a fair comparison.
 We observe that our method achieves an average ADE/FDE of $7.61$/$14.30$ in pixel coordinate, which achieves the best performance among all the current methods, regardless of the involvement of image data.
Specifically, our MID outperforms the current state-of-the-art T+I method Y-Net+TTST on the ADE metric.
Note that our method did not use the image data and apply any post-processing such as the Test-Time Sampling Trick (TTST)~\cite{mangalam2021goals}. We provide the results with the sampling trick in the supplementary.

\begin{figure}[t]
	\centering
	\includegraphics[width=1.0\linewidth]{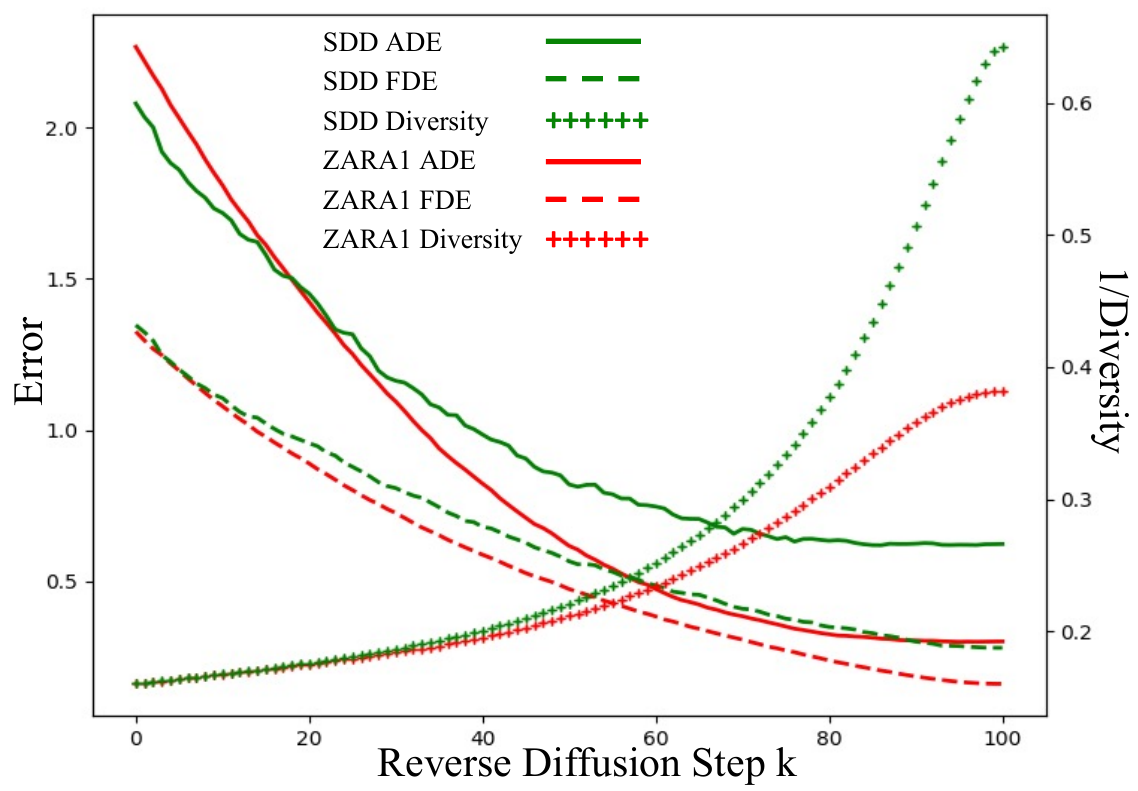}
	\caption{Trade-off between determinacy (ADE/FDE) and diversity within reverse diffusion steps from 0 to 100.}
	\label{fig:trade_off}
	\vspace{-0.6cm}
\end{figure}

\begin{figure*}[h]
    \scriptsize
    \setlength{\tabcolsep}{1.5pt}
    \centering
    \includegraphics[width=1\linewidth]{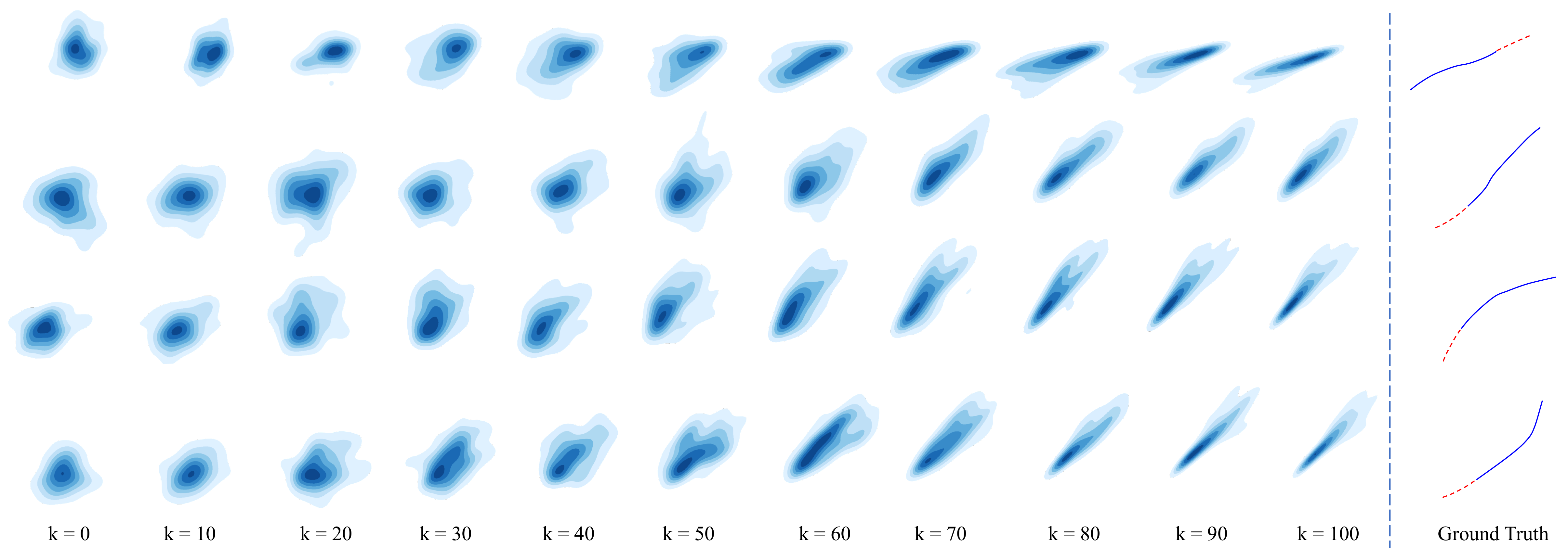}\\
    \caption{Visualization of generated trajectories at each diffusion time step $t$. We can see that the reverse diffusion process progressively reduces the indeterminacy and reaches the desired trajectory through the time step. Starting from a normal distribution at $t=0$ that corresponds to all walkable areas, and the observed paths (red dash), our MID method successfully eliminates improbable trajectories and gradually fits the ground truth future paths (blue line). Best viewed in color.}
    \label{fig:qualitative_diffuse}
    \vspace{-0.2cm}
\end{figure*}

We also conducted experiments on the ETH-UCY dataset and tabulated the results in Table~\ref{table:eth_ucy}.
Our method achieves a comparable performance with only trajectory input in 20 sampling, with an average performance of $0.21$ ADE and $0.38 $ FDE. We
found that MID are benefited more on the larger dataset (e.g. the SDD dataset).

\subsection{Ablation Studies}
In this subsection, we conducted ablation studies to investigate the effectiveness of each key component including diffusion model and Transformer architecture. Then, we provided a detailed analysis for reverse diffusion process.

\textbf{Diffusion Model:}
In order to examine the importance of our diffusion model, we degraded our MID into a CVAE based framework, Trajectron++.
We replaced the decoder from commonly-used LSTM to our Transformer in this CVAE based framework to verify whether the performance boost comes from the Transformer.
The group 2 and 4 in Table~\ref{table:ablation_ddpm} show the performance comparison.
We observe that using the same encoder and decoder but without our diffusion model, the results degrade significantly, demonstrating the effectiveness of our diffusion model.
Besides, only replacing the decoder with our Transformer architecture in the CVAE based framework does not improve performance as shown in group 4 in Table~\ref{table:ablation_ddpm}.

\textbf{Transformer Architecture:} We also conducted experiments on the decoder architecture of MID. According to group 1 and 3 in Table~\ref{table:ablation_ddpm}, Transformer outperforms the Linear and LSTM architecture by a large margin. It indicates the Transformer architecture is effective for MID to model the temporal dependencies of trajectory. Besides, we evaluated the Transformer architectures with different dimensions. As tabulated in Table~\ref{table:ablation_ddpm} group 1 and 2, we observe that the Transformer with 512 dimensions leads to the best performance, and further increasing the Transformer dimension or model parameters does not yield better results.

\textbf{Analysis of Reverse Diffusion Process:}
To further explore the reverse diffusion process, we generated $20$ trajectories at each reverse diffusion step and analyzed the gradual change of the distribution.
We provide an analysis between the reverse diffusion step and the corresponding diversity and ADE/FDE, as illustrated in Figure~\ref{fig:trade_off}. The trajectory diversity is calculated as the average of Euclidean distance between any of the two in the generated $20$ trajectories. When the reverse diffusion step is small, the trajectory distribution is more indeterminate and produces highly diverse trajectories. As the reverse diffusion step increases, we observe the decline in diversity and the rise of determinacy. With our MID framework, we can control the degree of indeterminacy by adjusting the step numbers, and achieve a flexible trade-off between the diversity and determinacy of the generated trajectories.

In addition, we visualize the distribution of trajectories as contours in Figure~\ref{fig:qualitative_diffuse} and each contour map is sampled by ten steps interval. We see that the contours are diverse at the early stage of the diffusion process, and deform gradually to be more concentrated and fit to the ground truth trajectory.

\begin{figure*}[h]
    \scriptsize
    \setlength{\tabcolsep}{1.5pt}
    \centering
    \includegraphics[width=1\linewidth]{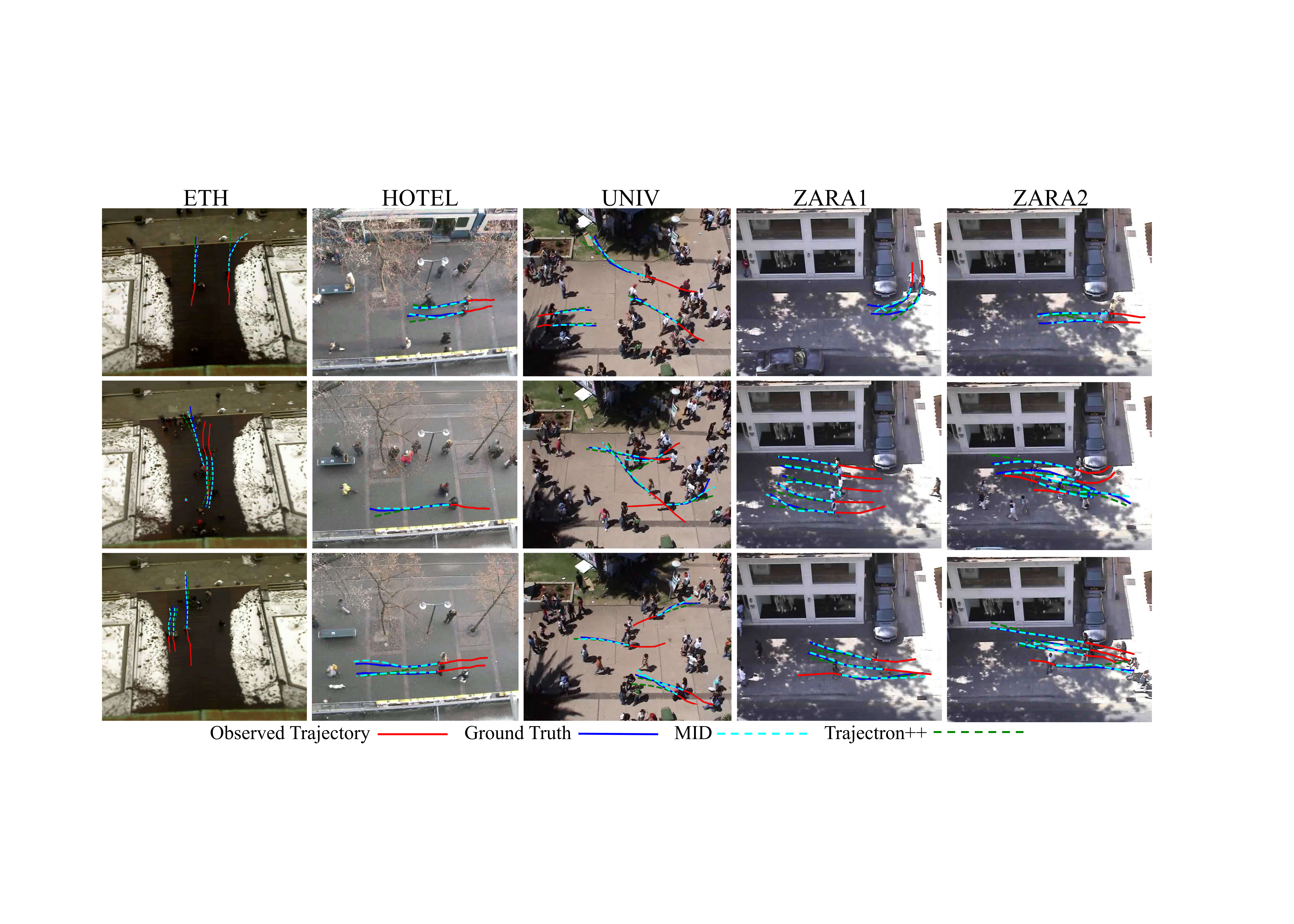}\\
    \vspace{-3pt}
    \caption{Visualization of predicted trajectories on the ETH/UCY Dataset. Given the observed trajectories (red), we illustrate the ground truth paths (blue) and predicted trajectories by MID (dashed cyan) and Trajectron++ (green) for five different scenes. We see that our results are much closer to the ground truth compared with Trajectron++. Best viewed in color and zoom-in for more clarity.}
    \label{fig:qualitative_eth_ucy}
    \vspace{-0.3cm}
\end{figure*}

\begin{figure*}[h]
    \scriptsize
    \setlength{\tabcolsep}{1.5pt}
    \centering
    \includegraphics[width=1\linewidth]{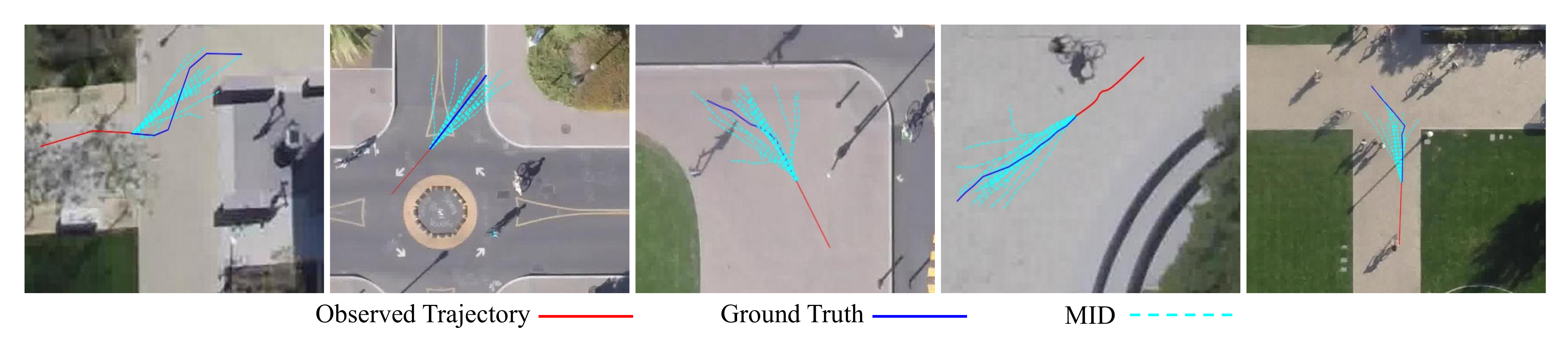}\\
    \vspace{-3pt}
    \caption{Visualization of generated trajectories in the Stanford Drone Dataset. Given the observed trajectories (red), we illustrate the ground truth paths (blue) and predicted best-of-20 trajectories by MID (dashed cyan) in different scenes. The blue line is covered by the cyan dashes in all scenes.
    Best viewed in color and zoom-in for more clarity.}
    \label{fig:qualitative_sdd}
    \vspace{-0.5cm}
\end{figure*}

\subsection{Qualitative Evaluation}
We further investigated the ability of our framework by the qualitative results.
Figure~\ref{fig:qualitative_eth_ucy} illustrates the most-likely predictions of our MID and Trajectron++~\cite{salzmann2020trajectron++} on all five scenes on the ETH/UCY dataset. The qualitative results show that both MID and Trajectron++ fits the ground truth paths well. We observe that Trajectron++ performs similarly to MID for short-term forecasting yet a little deviates from the ground truth path for longer prediction. Besides, we visualize multiple predicted trajectories on SDD in Figure~\ref{fig:qualitative_sdd}.
We observe that all predictions show their feasibility conditioned on the observed trajectories. Though reducing the ambiguity with the reverse diffusion model, We found that generated trajectories are still full of the diversity in a walkable region.

\section{Conclusion \& Discussion}
In this paper, we introduced a new MID framework to formulate trajectory prediction with motion indeterminacy diffusion. In this framework, we learned a parameterized Markov chain conditioned on the observed trajectories to gradually discard the indeterminacy from ambiguous areas to acceptable trajectories.
By adjusting the length of the chain, we can achieve the trade-off between diversity and determinacy. Besides, we designed a Transformer-based architecture as the core network of our method to model complex temporal dependency in trajectories.
Experimental results demonstrate the superiority of our method which achieves state-of-the-art performance on the Stanford Drone and ETH/UCY benchmarks.

\textbf{Broader Impact:} MID could be applied to a wide range of applications with human-robots interaction. With indeterminacy modeling, we can generate accurate and reasonable future trajectories, which helps much with decision making in auto-driving. Besides, MID can adjust the degree of indeterminacy, which has the potential to be applied in dynamic and interactive environments.

\textbf{Limitations:} Despite the promising performance and an applicable trade-off nature, the time cost at reverse diffusion process could be expensive due to multiple steps (100 steps in our experiments). When evaluated with 512 trajectories on the ZARA1 dataset, Trajectron++ needs $\mathbf{0.443s}$ but MID will need $\mathbf{17.368s}$ with 100 diffusion steps setting.
Fortunately, many recent efforts have been made to significantly reduce the sampling cost while keeping the high generation performance ~\cite{nichol2021improved,song2020score,jolicoeurmartineau2021gotta,san2021noise,watson2021learning}. However, plugging these methods in our MID is not trivial.  We leave it as future work to build a more efficient system.

\noindent \textbf{Acknowledgments} This work was supported in part by the National Natural Science Foundation of China under Grant 62125603, and Grant U1813218, in part by a grant from the Beijing Academy of Artificial Intelligence (BAAI).

{\small
\bibliographystyle{ieee_fullname}
\bibliography{mid}
}

\clearpage
\begin{appendix}

\section{Detailed Derivations}

\subsection{Derivations of Loss Function}
We give the derivations to obtain our loss function as:
\begin{equation}
\setlength{\abovedisplayskip}{1pt}
\setlength{\belowdisplayskip}{1pt}
\begin{aligned}
\label{eq: loss_1}
L(\theta,\psi) & = \mathbb{E}_{q}[ \sum_{k=1}^{K}-\log\frac{p_{\theta}(\mathbf{y}_{k-1} | \mathbf{y}_{k}, \mathbf{f})}{q(\mathbf{y}_{k}|\mathbf{y}_{k-1})}] \\
&= -\mathbb{E}_{q}[ \sum_{k=2}^{K}\log\frac{p_{\theta}(\mathbf{y}_{k-1} | \mathbf{y}_{k}, \mathbf{f})}{q(\mathbf{y}_{k}|\mathbf{y}_{k-1})} + \log\frac{p_{\theta}(\mathbf{y}_{0} | \mathbf{y}_{1}, \mathbf{f})}{q(\mathbf{y}_{1}|\mathbf{y}_{0})} ] \\
&= -\mathbb{E}_{q}[ \sum_{k=2}^{K}\log {p_{\theta}(\mathbf{y}_{k-1} | \mathbf{y}_{k}, \mathbf{f})} \frac{q(\mathbf{y}_{k-1}|\mathbf{y}_{0})}{ q(\mathbf{y}_{k-1}|\mathbf{y}_{k},\mathbf{y}_{0} ) q(\mathbf{y}_{k}|\mathbf{y}_{0})  }\\
&+ \log\frac{p_{\theta}(\mathbf{y}_{0} | \mathbf{y}_{1}, \mathbf{f})}{q(\mathbf{y}_{1}|\mathbf{y}_{0})} ]\\
&= -\mathbb{E}_{q}[ \sum_{k=2}^{K}\log\frac {p_{\theta}(\mathbf{y}_{k-1} | \mathbf{y}_{k}, \mathbf{f})}{q(\mathbf{y}_{k-1}|\mathbf{y}_{k},\mathbf{y}_{0})} + \sum_{k=2}^{K}\log\frac {q(\mathbf{y}_{k-1}|\mathbf{y}_{0})}{q(\mathbf{y}_{k}|\mathbf{y}_{0})}
\\
&+\log\frac{p_{\theta}(\mathbf{y}_{0} | \mathbf{y}_{1}, \mathbf{f})}{q(\mathbf{y}_{1}|\mathbf{y}_{0})} ]\\
&= \mathbb{E}_{q}[ \sum_{k=2}^{K}-\log\frac {p_{\theta}(\mathbf{y}_{k-1} | \mathbf{y}_{k}, \mathbf{f})}{q(\mathbf{y}_{k-1}|\mathbf{y}_{k},\mathbf{y}_{0})}
\\
&- \log{p_{\theta}(\mathbf{y}_{0} | \mathbf{y}_{1}, \mathbf{f})}  + \log q(\mathbf{y}_{K}|\mathbf{y}_{0})].\\
\end{aligned}
\end{equation}
We ignore the last term because it has no learnable parameters and get the loss function as:
\begin{equation}
\setlength{\abovedisplayskip}{1pt}
\setlength{\belowdisplayskip}{1pt}
\begin{aligned}
\label{eq: loss_paper} L(\theta,\psi) & =  \mathbb{E}_{q}[\sum_{k=2}^{K}D_{KL}(q(\mathbf{y}_{k-1}\vert \mathbf{y}_{k},\mathbf{y}_{0}) \Vert p_{\theta}(\mathbf{y}_{k-1} \vert \mathbf{y}_{k}, \mathbf{f})) \\
&  -  \log p_{\theta}(\mathbf{y}_{0} \vert \mathbf{y}_{1},\mathbf{f})].
\end{aligned}
\end{equation}

\subsection{Derivations of Reparameterization}
As shown in the loss function, we should match the reverse transition $p_{\theta}(\mathbf{y}_{k-1} \vert \mathbf{y}_{k}, \mathbf{f}) $ and the ground-truth $q(\mathbf{y}_{k-1}\vert \mathbf{y}_{k},\mathbf{y}_{0})$, both of which are in Gaussian. We can convert the KL divergence of two Gaussian distributions as the difference of the means.
We calculate the mean of posterior in a closed form:
\begin{equation}
\setlength{\abovedisplayskip}{1pt}
\setlength{\belowdisplayskip}{1pt}
\begin{aligned}
\label{eq: q_mu_sigma}  \Tilde{\boldsymbol{\mu}}_{k}(\mathbf{y}_{k},\mathbf{y}_{0}) &= \frac{\sqrt{\Bar{\alpha}_{k-1}}\beta_{k}}{1-\Bar{\alpha}_{k}}\mathbf{y}_{0} + \frac{\sqrt{\alpha_{k}}(1-\Bar{\alpha}_{k-1})}{1-\Bar{\alpha}_{k}}\mathbf{y}_{k},
\end{aligned}
\end{equation}
where $\alpha_{k} = 1 - \beta_{k}$ and $\Bar{\alpha}_{k} = \prod_{s=1}^{k}\alpha_{s}$.
By the reparameterization, we formulate the $\mathbf{y}_{k}$ as a function of $\mathbf{y}_{0}$ and $\epsilon$:
\begin{equation}
\begin{aligned}
\label{eq: reparameterization}
\mathbf{y}_{k}(\mathbf{y}_{0},\epsilon ) = \sqrt{\Bar{\alpha}_{k}} \mathbf{y}_{0} + \sqrt{1-\Bar{\alpha}_{t}} \epsilon,
\end{aligned}
\end{equation}
where $\epsilon \sim \mathcal{N}(0,\mathbf{I})$ is a random variable, and we have
\begin{equation}
\begin{aligned}
\label{eq: y0}
\mathbf{y}_{0}= \frac{1}{\sqrt{\Bar{\alpha}_{k}}}( \mathbf{y}_{k}(\mathbf{y}_{0},\epsilon) - \sqrt{1-\Bar{\alpha}_{k}}\epsilon).
\end{aligned}
\end{equation}
Then we reformulate $\Tilde{\boldsymbol{\mu}}_{k}(\mathbf{y}_{k},\mathbf{y}_{0}) $:
\begin{equation}
\setlength{\abovedisplayskip}{1pt}
\setlength{\belowdisplayskip}{1pt}
\begin{aligned}
\label{eq: q_mu_sigma2}  \Tilde{\boldsymbol{\mu}}_{k}(\mathbf{y}_{k}(\mathbf{y}_{0},\epsilon),\epsilon) &=(
\frac{\sqrt{\Bar{\alpha}_{k-1}}\beta_{k}}{\sqrt{\Bar{\alpha}_{k}}(1-\Bar{\alpha}_{k})}
+\frac{\sqrt{\alpha_{k}}(1-\Bar{\alpha}_{k-1})}{1-\Bar{\alpha}_{k}}
 )\mathbf{y}_{k}(\mathbf{y}_{0},\epsilon) \\
& + \frac{\sqrt{1-\Bar{\alpha}_{k}}\sqrt{\Bar{\alpha}_{k-1}}\beta_{k}}{\sqrt{\Bar{\alpha_{k}}}(1-\Bar{\alpha}_{k})}
 \epsilon \\
& =\frac{\sqrt{\Bar{\alpha}_{k-1}}\beta_{k}+ \sqrt{\Bar{\alpha}_{k}} \sqrt{\alpha_{k}}(1-\Bar{\alpha}_{k-1}) }  {\sqrt{\Bar{\alpha_{k}}}(1-\Bar{\alpha}_{k})} \mathbf{y}_{k}(\mathbf{y}_{0},\epsilon) \\
& - \frac{\beta_{k}}{\sqrt{\alpha_{k}}\sqrt{1-\Bar{\alpha}_{k}}}
 \epsilon \\
 & =\frac{\sqrt{\Bar{\alpha}_{k-1}}( \beta_{k}+ \alpha_{k}(1-\Bar{\alpha}_{k-1})) }
 {\sqrt{\Bar{\alpha_{k}}}(1-\Bar{\alpha}_{k})}\mathbf{y}_{k}(\mathbf{y}_{0},\epsilon) \\
& - \frac{\beta_{k}}{\sqrt{\alpha_{k}}\sqrt{1-\Bar{\alpha}_{k}}}
 \epsilon \\
  & = \frac{1}  {\sqrt{\alpha_{k}}}
  \mathbf{y}_{k}(\mathbf{y}_{0},\epsilon) - \frac{\beta_{k}}{\sqrt{\alpha_{k}}\sqrt{1-\Bar{\alpha}_{k}}}
 \epsilon \\
 & = \frac{1}  {\sqrt{\alpha_{k}}}(
  \mathbf{y}_{k}(\mathbf{y}_{0},\epsilon) - \frac{\beta_{k}}{\sqrt{1-\Bar{\alpha}_{k}}}
 \epsilon) .
\end{aligned}
\end{equation}
Therefore, the $D_{KL}$ can be formulated as:
\begin{equation}
\begin{aligned}
\label{eq: kl_mu}  D_{KL} & \!=\! \mathbb{E}_{\mathbf{y}_{0},\epsilon}\left[ \lambda \Vert \Tilde{\boldsymbol{\mu}}_{k}(\mathbf{y}_{k}(\mathbf{y}_{0},\epsilon),\epsilon) \!-\! \boldsymbol{\mu}_{\theta}(\mathbf{y}_{k},k,\mathbf{f})\Vert^{2} \right] \\
&\!=\! \mathbb{E}_{\mathbf{y}_{0},\epsilon}\left[ \lambda \big\Vert \frac{1}  {\sqrt{\alpha_{k}}}(
  \mathbf{y}_{k}(\mathbf{y}_{0},\epsilon) \!-\! \frac{\beta_{k}}{\sqrt{1-\Bar{\alpha}_{k}}}
 \epsilon) \!-\! \boldsymbol{\mu}_{\theta}(\mathbf{y}_{k},k,\mathbf{f}) \big\Vert^{2} \right],
 \end{aligned}
\end{equation}

\algrenewcommand\algorithmicindent{0.5em}%
\begin{figure}[t]
\begin{minipage}[t]{0.495\textwidth}
\begin{algorithm}[H]
  \caption{Pseudocode for MID Training Procedure}
  \small
  \begin{algorithmic}[1]
    \Repeat
      \State Sample trajectory $(\bx,\by) \sim q_{\text{data}}$
      \State $\bx:$ Observed Trajectory
      \State $\by:$ Future Trajectory
      \State $\by_0 = \by$
      \State $k \sim \mathrm{Uniform}(\{1, \dotsc, K\})$
      \State $\epsilon\sim\mathcal{N}(\mathbf{0},\mathbf{I})$
      \State Take gradient descent step on
      \Statex $\qquad \nabla_{(\theta,\psi)} \left\| \epsilon - \epsilon_{(\theta,\psi)}(\sqrt{\bar\alpha_k} \by_0 + \sqrt{1-\bar\alpha_k}\epsilon, k, \bx) \right\|^2$
    \Until{converged}
  \end{algorithmic}
\label{algorithm:train}
\end{algorithm}
\end{minipage}
\hfill

\begin{minipage}[t]{0.495\textwidth}
\begin{algorithm}[H]
  \caption{Pseudocode for MID Sampling Procedure}
  \small
  \begin{algorithmic}[1]
    \vspace{.04in}
    \State \textbf{Input:} Observed Trajectory $\bx$
    \State \textbf{Output:} Predicted Trajectory $\by$
    \State Sample $\by_K \sim \mathcal{N}(\mathbf{0}, \mathbf{I})$
    \For{$k=K, \dotsc, 1$}
      \State $\mathbf{z} \sim \mathcal{N}(\mathbf{0}, \mathbf{I})$ if $k > 1$, else $\mathbf{z} = \mathbf{0}$
      \State $\by_{k-1} = \frac{1}{\sqrt{\alpha_k}}\left( \by_k - \frac{\beta_k}{\sqrt{1-\bar\alpha_k}} \epsilon_{(\theta,\psi)}(\by_k, k,\bx) \right) + \sqrt{\beta_k} \mathbf{z}$
    \EndFor
    \State $\by = \by_{0}$
    \State \textbf{return} $\by$
    \vspace{.04in}
  \end{algorithmic}
\label{algorithm:sample}
\end{algorithm}
\end{minipage}
\end{figure}

\begin{figure*}[h]
\vspace{-0.2cm}
  \centering
  \includegraphics[width=8cm]{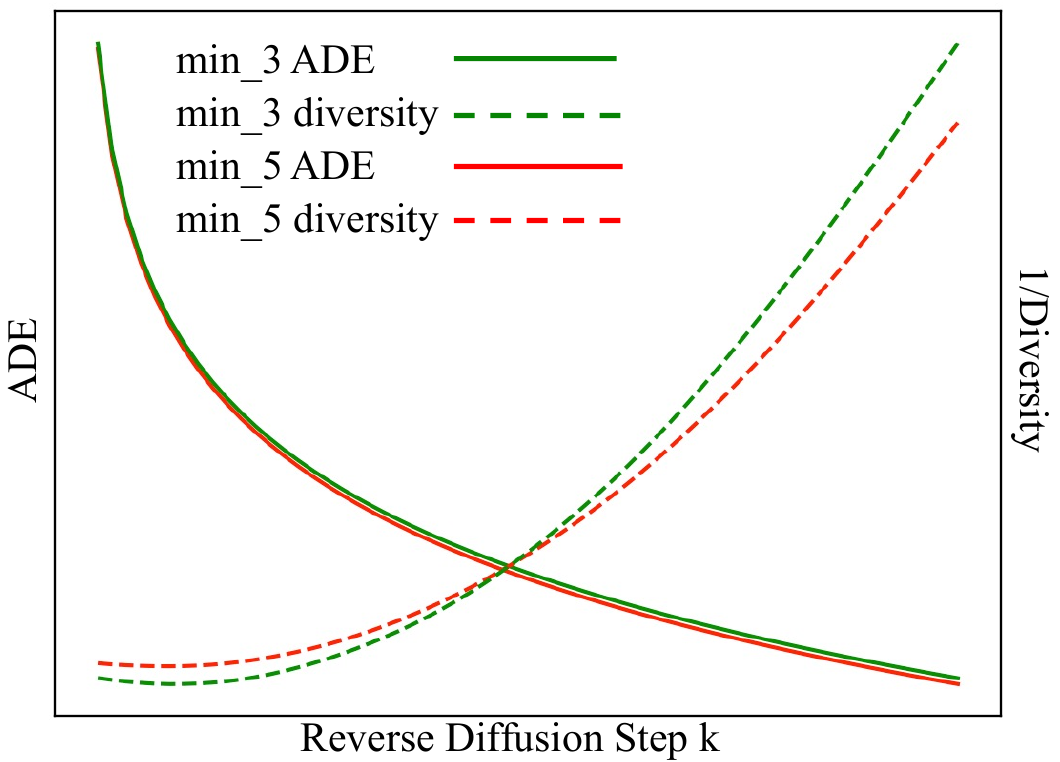}
  \hspace{0.01in}
  \includegraphics[width=8cm]{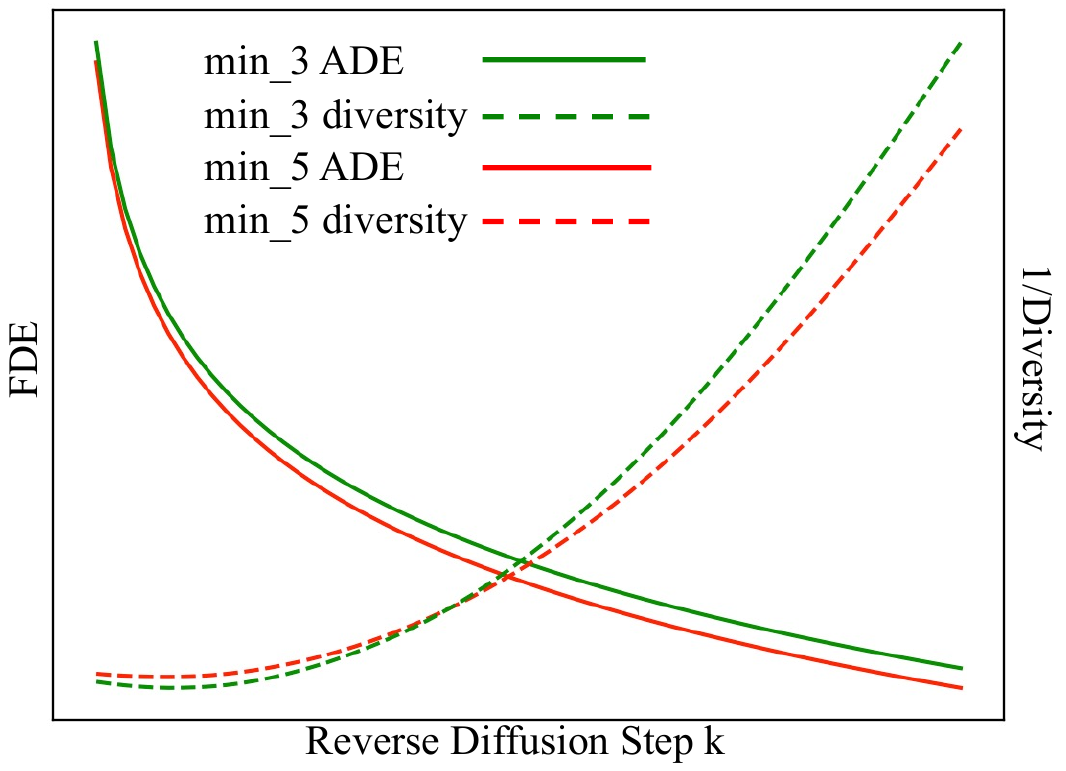}
  \vspace{-0.1cm}
  \caption{Trade-off between min$\_3$/min$\_5$ and diversity.}
  \label{fig: min3_min5}
  \vspace{-0.3cm}
\end{figure*}

Then we show why the last term $-  \log p_{\theta}(\mathbf{y}_{0} \vert \mathbf{y}_{1},\mathbf{f})$  is tractable with the same formulation form of $D_{KL}$ at $k=1$. The term $-  \log p_{\theta}(\mathbf{y}_{0} \vert \mathbf{y}_{1},\mathbf{f})$ means that the outputs of prediction model should follow the distribution of real data. Considering the reverse transition $p_{\theta}(\mathbf{y}_{0} \vert \mathbf{y}_{1},\mathbf{f}) $ is Gaussian, we also revert this loss as the difference between the mean of Gaussian transition $\boldsymbol{\mu}_{\theta}(\mathbf{y}_{k},k,\mathbf{f})$ and the ground truth $\mathbf{y}_{0}$ as $\mathbb{E}\left[ \lambda \Vert \mathbf{y}_{0} - \boldsymbol{\mu}_{\theta}(\mathbf{y}_{1},1,\mathbf{f})\Vert^{2} \right] $. Moreover, for the $D_{KL}$ under $k=1$, we have
\begin{equation}
\begin{aligned}
\label{eq: l0}
\Tilde{\boldsymbol{\mu}}_{1}(\mathbf{y}_{1}(\mathbf{y}_{0},\epsilon),\epsilon) &=
\frac{1}  {\sqrt{\alpha_{1}}}(
  \mathbf{y}_{1}(\mathbf{y}_{0},\epsilon) - \frac{\beta_{1}}{\sqrt{1-\Bar{\alpha}_{1}}}
 \epsilon) \\
 & = \frac{1}  {\sqrt{\Bar{\alpha}_{1}}}(
  \mathbf{y}_{1}(\mathbf{y}_{0},\epsilon) - \sqrt{1-\Bar{\alpha}_{1}}\epsilon).
\end{aligned}
\end{equation}
With~\eqref{eq: y0}, we get $ \Tilde{\boldsymbol{\mu}}_{1}(\mathbf{y}_{1}(\mathbf{y}_{0},\epsilon),\epsilon) = \mathbf{y}_{0}$, which demonstrates the of losses with both $k=1$ and $k \geq 2$ are in the same form.

As shown in~\eqref{eq: q_mu_sigma2} and \eqref{eq: kl_mu}, the loss function expects the model to predict $\frac{1}  {\sqrt{\alpha_{k}}}(
  \mathbf{y}_{k}(\mathbf{y}_{0},\epsilon) - \frac{\beta_{k}}{\sqrt{1-\Bar{\alpha}_{k}}}
 \epsilon)$ given the inputs $\mathbf{y}_{k}(\mathbf{y}_{0},\epsilon)$ and $\mathbf{f} $. Since the $\mathbf{y}_{k}(\mathbf{y}_{0},\epsilon)$ is the input, we only need a network to predict $\epsilon $ as  $\epsilon_{\theta}(\mathbf{y}_{k}(\mathbf{y}_{0},\epsilon),k,\mathbf{f} ) $. Thus, the final loss function is formulated as:
 \begin{equation}
\begin{aligned}
\label{eq: loss_simple} L(\theta, \psi) = \mathbb{E}_{\epsilon,\mathbf{y}_{0},k} \Vert \epsilon - \epsilon_{(\theta, \psi)}(\mathbf{y}_{k}, k, \mathbf{x})  \Vert,
\end{aligned}
\end{equation}
where $\psi $ denotes we further consider the encoder network in the loss function. Once the network $\epsilon_{(\theta, \psi)}(\mathbf{y}_{k}, k, \mathbf{x}) $ is trained, we can use this network to obtain the mean of  Gaussian transition.
 \begin{equation}
\begin{aligned}
\label{eq: samping}
 \boldsymbol{\mu}_{\theta}(\mathbf{y}_{k},k,\mathbf{f})=
\frac{1}  {\sqrt{\alpha_{k}}}(
  \mathbf{y}_{k}(\mathbf{y}_{0},\epsilon) - \frac{\beta_{k}}{\sqrt{1-\Bar{\alpha}_{k}}}
 \epsilon_{(\theta, \psi)}(\mathbf{y}_{k}, k, \mathbf{x})).
\end{aligned}
\end{equation}
Furthermore, the trajectory in next step is predicted as:
 \begin{equation}
\begin{aligned}
\label{eq: samping2}
 \mathbf{y}_{k-1}=
\frac{1}  {\sqrt{\alpha_{k}}}(
  \mathbf{y}_{k}(\mathbf{y}_{0},\epsilon) - \frac{\beta_{k}}{\sqrt{1-\Bar{\alpha}_{k}}}
 \epsilon_{(\theta, \psi)}(\mathbf{y}_{k}, k, \mathbf{x})) + \sqrt{\beta_{k}} \mathbf{z},
\end{aligned}
\end{equation}
where $\mathbf{z} \sim \mathcal{N}(\mathbf{0},\mathbf{I})$.

\section{Implementation Details}

In this section, we introduce the implementation details of our method, including the hyper-parameters for training, the network architecture, the algorithms of training and inference, and the attached code.

\textbf{Diffusion Process and Hyper-parameters:} We set the lower bound of variance scheduler $\beta_{1}$ to 0.0001 and upper bound $\beta_{K}$ to 0.05, and $\beta_{k}$ is uniformly sampled between the bounds. For the main Transformer network in diffusion model $\epsilon_{\theta}$, we devise three Transformer Encoder layers where each has the dimension of 512, feedforward dimension of 1024 and 4 attention heads. For the encoder $\mathcal{F}_{\psi}$, we utilize the default configuration provided by Trajectron++~\cite{salzmann2020trajectron++}.

\textbf{Upsample-Downsample Layers:} We employ a MLP-based sub-network to upsample the raw trajectory from 2d to 512d, and downsample the output of the Transformer such that 512d-256d-2d as the final output of the network. Each sub-network, denoted by $M$ and parameterized by ${\phi}$, contains three MLP layers which we can formulate as:
\begin{equation}
    M_{\phi}(\mathbf{h}, k, \mathbf{f}) = (\mathbf{W}_{1}\mathbf{h} + \mathbf{b}_{1}) \odot \sigma(\mathbf{W}_{2}\mathbf{c}+\mathbf{b}_{2}) + (\mathbf{W}_{3}\mathbf{c} + \mathbf{b}_{3}).
\end{equation}
$\mathbf{c}$ is the concatenation of step number embedding and state embedding such that $\mathbf{c} = [k, \sin(k),\cos(k), \mathbf{f}]$ and
$\mathbf{h}$ denotes the input trajectory feature of the sub-network. $\mathbf{W}_1$, $\mathbf{W}_2$, $\mathbf{W}_3$ and $\mathbf{b}_1$, $\mathbf{b}_2$, $\mathbf{b}_3$ are the trainable parameters of the MLP layers. $\sigma$ corresponds to a sigmoid function.

\begin{table}
\caption{Ablation studies on sampling number on SDD.}
\vspace{-0.4cm}
\linespread{3.0}
\renewcommand\arraystretch{1.1}
\renewcommand\tabcolsep{3.7pt}
\begin{center}
\newcolumntype{g}{>{\columncolor{Gray}}c}
\newcolumntype{y}{>{\columncolor{LightCyan}}c}
\newcolumntype{d}{>{\columncolor{DarkCyan}}c}

\begin{tabular}{ c | c|g }
\hline
\hline
\textbf{Sampling}  &\textbf{ADE} & \textbf{FDE} \\
    \hline
20 & 7.61 & 14.30 \\
40 & 6.84 & 12.00 \\
20$\times$20 & 5.42 & 9.47 \\
\hline\hline
\end{tabular}
\end{center}
\label{table:ablation_sampling}
\vspace{-0.3cm}
\end{table}

\textbf{Reproducibility:}
For better understanding and reproduction, we provide Algorithm~\ref{algorithm:train} and Algorithm~\ref{algorithm:sample} showing the training and inference procedure of our MID framework. Furthermore,
the code can be found in \url{https://github.com/gutianpei/MID}.

\section{Additional Experiments}
We also respectively report the ADE (left) and FDE (right) curves of min\_3/min\_5 metrics within reverse diffusion steps from 0 to 100 in Figure~\ref{fig: min3_min5}.
We can observe reducing the diversity also leads to better predictions with fewer samples, which demonstrates that diversity and determinacy are still contradictory with few samples.

Additionally, we found that the sampling trick is very effective to improve model performance. Sampling tricks usually add the number of sampling and do the post-processing (clustering in YNet~\cite{mangalam2021goals} and choosing best in Expert~\cite{zhao2021you}). As shown in Table~\ref{table:ablation_sampling}, the performance is improved significantly when we add the number of sampling like Expert. However, we don't encourage to use more samplings since more samplings indicate more computation cost.

\end{appendix}
\end{document}